\documentclass{article}

\PassOptionsToPackage{square,sort,comma,numbers}{natbib}
\usepackage[final]{nips_2018}

\usepackage[utf8]{inputenc} 
\usepackage[T1]{fontenc}    
\usepackage{url}            
\usepackage{booktabs}       
\usepackage{amsfonts}       
\usepackage{nicefrac}       
\usepackage{microtype}      
\usepackage{amssymb} 
\usepackage{amsmath} 
\usepackage[inline]{enumitem}
\usepackage{multicol}
\usepackage[usenames]{color} 
\usepackage{graphicx}
\usepackage{acronym}
\usepackage{todonotes}
\usepackage{censor}
\usepackage{appendix}
\usepackage{cleveref}
\crefname{appsec}{Appendix}{Appendices}

\usepackage[compact]{titlesec}
\titlespacing*{\subsection}{0pt}{1ex}{0ex}
\titlespacing*{\paragraph}{0pt}{1ex}{1ex}

\acrodef{AWS}{Amazon Web Services}
\acrodef{ML}{Machine Learning}
\acrodef{RL}{Reinforcement Learning}
\acrodef{GDPR}{General Data Protection Regulation}
\acrodef{MLAAS}{\ac{ML}-as-a-service}
\acrodef{GPU}{Graphics Processing Unit}
\acrodef{ROC}{Receiver Operating Characteristic}
\acrodef{PR}{Precision Recall}
\acrodef{ETL}{Extract/Transform/Load}
\acrodef{MAD}{Median Absolute Deviation}
\acrodef{MCD}{Minimum Covariance Determinant}
\acrodef{ADR}{Adaptable Damped Reservoir}
\acrodef{PCA}{Principal Components Analysis}
\acrodef{PSI}{Population Stability Index}
\acrodef{KL}{Kullback-Leibler}
\acrodef{KLIEP}{\ac{KL} Importance Estimation Procedure}
\acrodef{SVM}{Support Vector Machine}
\acrodef{EDDM}{Early Drift Detection Method}
\acrodef{HPO}{Hyper-Parameter Optimization}
\acrodef{BO}{Bayesian Optimization}
\acrodef{SOTA}{State-of-the-Art}

\newcommand{\eg}{\emph{e.g.}}
\newcommand{\ie}{\emph{i.e.}}
\newcommand{\etc}{\emph{etc}}

\Crefname{figure}{Fig.}{Figs.}
\Crefname{section}{Sec.}{Secs.}

\title{Continual Learning in Practice}

\author{
  Tom Diethe, Tom Borchert, Eno Thereska, Borja Balle, Neil Lawrence\\
  \texttt{\{tdiethe,borcht,enother,pigem,lawrennd\}@amazon.com}
}

\begin{document}

\maketitle

\begin{abstract}
  
This paper describes 
a reference architecture for self-maintaining systems that can learn continually, as data arrives. In environments where data evolves, we need architectures that  manage \ac{ML} models in production, adapt to shifting data distributions, cope with outliers, retrain when necessary, and adapt to new tasks. This represents continual AutoML or Automatically Adaptive Machine Learning. We describe the challenges and proposes a reference architecture.

\end{abstract}

\section{Introduction}

\acf{ML} technologies have been widely adopted by industry, but were developed mainly in academia. The principle characteristic of a \ac{ML} system is that it is \emph{data-driven}. A \ac{ML} system is a software system where the software is written by data.
Our current approach to model deployment is developed around the classical view of \ac{ML}. A model is \emph{trained} on a training set and then \emph{deployed} into production as part of a wider software \emph{ecosystem}. ML deployment strategies leverage our modern understanding of software engineering and continuous deployment. From that perspective, a trained \ac{ML} model is merely another software component that can be deployed through existing pipelines. In such pipelines software is verified at the moment of deployment by a battery of tests that ensure the new component is 
(at the very least) not degrading system performance.

This \emph{de-facto} standard for model deployment has a critical flaw: data is not static, it evolves. As a result, verification performed during deployment becomes invalid through the passage of time.
Continual learning aims to address this challenge by continuous update of deployed models. Our paper presents the complementary perspective of the systems engineer. From a practical perspective, how should we validate and verify our \ac{ML} components in this evolving environment?


The modern approach systems engineering advocates strict boundaries and the use of micro-services, components, and interfaces. Extensive use of testing at the component and system level: unit tests (component) integration tests (system), regression tests (system) and acceptance tests (system) ensure that each component and the entire end-to-end workflow is not degraded through new deployments.
The first observation is that in an environment with continually evolving data, some, or all, of these tests need to be rerun. By analogy with \emph{regression} testing, we can think of this need as a need for \emph{progression} testing (see \Cref{sec:drift}).

As well as the challenge of continually evolving data, \ac{ML} systems erode component boundaries: 
\begin{enumerate*}[label=(\roman*)]
  \item system outputs may depend on the inputs in non-trivial 
ways, making integration testing challenging;
  \item \ac{ML} systems are often \emph{entangled} 
with control parameters that are interrelated, and external dependencies; 
  \item The outputs of \ac{ML} models may be used as the inputs to downstream systems, which themselves may involve \ac{ML} models \etc{}, which can result in \emph{hidden feedback loops}.
\end{enumerate*}

\ac{ML} models are compressors: taking the original input (training) data they produce compact representations (\eg{} neural network weights). In deployment new data is ingested, producing another compact representation: predictions. These representations are portable and easy to deploy in a software ecosystem. But entanglement means that even if downstream systems rely on sub-optimal models, it may become too risky to deploy the changes to the model, because the downstream system may be calibrated to the old model. 

\paragraph{Auto-Adaptive Machine Learning.}
The remedy we propose stems from a notion of `zero touch' \ac{ML}, i.e. \ac{ML} models that can be deployed and do not need to be maintained manually. Classical AutoML \cite{feurer2015efficient} speeds the creation of ML solutions. It does \emph{not} address the challenge of deployed ML systems we have outlined above. \ac{ML} requires a hypervisor: A system that monitors, creates and maintains deployed models. Our notion of Auto-Adaptive ML systems goes beyond AutoML by acknowledging the challenges of deployed ML systems.
Bringing about this vision requires the resolution of a chicken-and-egg problem: we cannot develop \ac{ML} methods to solve this challenge unless we have data from this challenge. In this paper we describe a reference architecture, one that we hope seeds research into a full solution of this challenge.



\section{Related Work}

Several companies have developed internal ML deployment environments. Facebook's ``FBLearner Flow'' \citep{fblearnerflow} eases the path to bringing \ac{ML} models into production while Uber's  \ac{MLAAS} platform ``Michelangelo'' \citep{michelangelo}, which 
\emph{``democratizes machine learning and makes scaling AI to meet the needs of business as easy as requesting a ride''}. 
But the motivation of both of these systems is smoothing the path to production. Our focus is maintenance of the model 
once deployment has occurred. Rapid deployment without continuous vigilance would exacerbate technical debt.
%
%
MacroBase \citep{bailis2017macrobase} is an analytic monitoring engine designed to prioritise human attention in large-scale datasets and data streams. MacroBase contains some similar notions, but is specialised for the task of finding and explaining unusual or interesting trends in data. 

\section{Reference Architecture}
\label{sec:architecture}

\begin{figure}[t]
  \centering
  \includegraphics[width=0.9\linewidth]{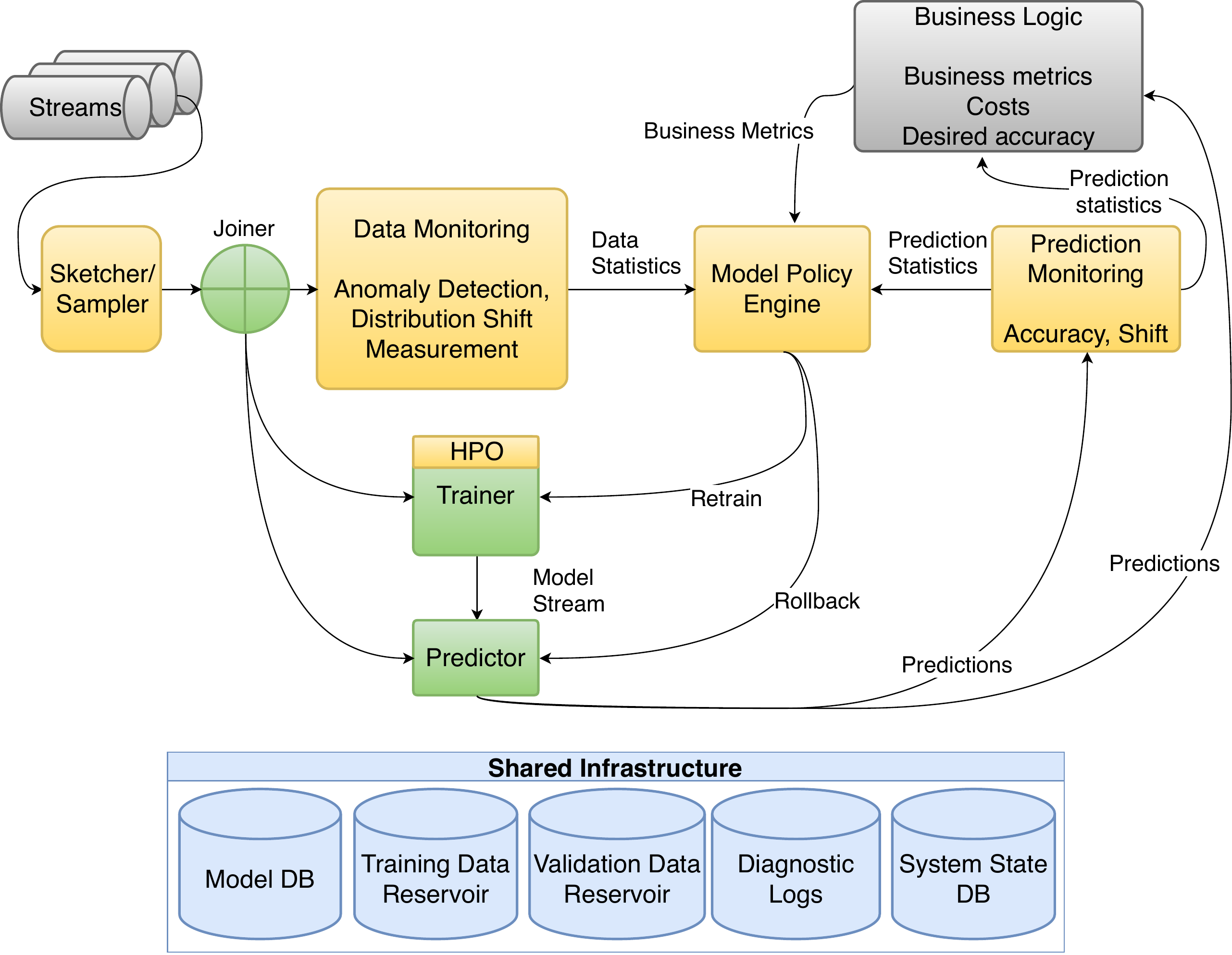}
  \caption{Data flow in the Auto-Adaptive Machine Learning architecture. See text for details.}
  \label{fig:architecture}
\end{figure}

Our vision is for Auto-Adaptive ML systems which rely on continual learning. But just as AutoML aims to automatically train any ML model, Auto-Adaptive ML should support any deployed \ac{ML} model. The focus for our reference architecture is classical learning methods as well as continual learning methods.
The overall system architecture is given in \Cref{fig:architecture}. Here we give a rationale for laying out the architecture as it is, and briefly state the purpose of each component. The following sub-sections will outline the role of each in more detail.

\paragraph{Streams:}  Both classical software and \ac{ML} systems process data in batches, but many modern software systems are based around \emph{streaming}. 
While it is possible to run a streaming system in batch mode the reverse is not true. This capability eases the transition for existing ML deployments. We build our ideas on top of streaming systems.
In a high data-rate streaming scenario, queries are never-ending, continuous, streaming queries. To deal with quantity of data we propose a sketcher (smart heavy down-sampler). 
In batch processing, data is there and you query it. In stream processing data might be arriving late, out of order, or it might be dropped altogether
: this is mitigated by the joiner 
component. 

\paragraph{Self-diagnosis through monitoring:} Auto-Adaptability implies that the system self-diagnoses when errors are likely to occur. The is enabled by the data monitoring subsystem 
which analyses the incoming streams looking for possible anomalies, drift, and change-points, and by the prediction monitoring subsystem 
, which analyses the prediction and health monitoring streams generated by the predictor subsystem
.

\paragraph{Self-correction policies:} The policy engine 
is responsible for updating models by triggering re-training or other actions. In its simplest form, the policy engine could use classical AutoML techniques in combination with retraining triggers. Decisions would be made on the basis of the data monitoring subsystem, the prediction monitoring subsystem, and any associated business logic. 

\paragraph{Self-management of resources:} The shared infrastructure 
can be seen as an intelligent cache that manages the storage requirements of the system.
%
There is the requirement to keep logs since customers might want to replay historical episodes. 

\setlength{\belowcaptionskip}{0pt}
\setlength{\textfloatsep}{10pt}

\subsection{Sketcher/Compressor}
\label{sec:sketch}
The sketcher/compressor component deals with the challenge that arises when data throughput is too high for the downstream training systems to cope with. At its simplest, the sketcher could simply down-sample, \eg{} uniformly or uniformly at random. The standard error for a sample size $s$ is proportional to $s^{-\frac{1}{2}}$, so \eg{} for 1000 records an error rate of $3\%$ can be expected. This serves as a baseline, but depending on the specific scenario, 
there are potentially much better options. We provide an overview of some sketching approaches in \Cref{sec:sota_sketch}.

\subsection{Joiner}
\label{sec:joiner}
In some cases labelled data is available directly with the incoming data streams, in which case a \emph{stream joiner} is unnecessary. In many cases, however, labels (or other forms of strong or weak feedback) required for training \ac{ML} models will be delayed or come from other sources. The joiner subsystem is required to ensure that the trainer and predictor receives the data in the format required. 

\subsection{Shared Infrastructure}
\label{sec:cache}
The shared infrastructure is shown as a collection of databases, but it is not assumed that the requirements of 
such a system are unbounded. 
This component contains the logic required to ensure optimal use of space, analogous to an operating system cache. 
Trained models (when feasible) are written to the \emph{Model DB} to ensure provenance (to be able to answer questions about why a particular decision was made at a particular time) and rollback functionality. The \emph{Training DB} holds a history of the training episodes from the Trainer subsystem. The \emph{Validation Data Reservoir} is used for validating models during training, and (subject to space constraints) can also be used for reproducing old models. We have made use of the \acl{ADR} as developed by \cite[Algo. 1]{bailis2017macrobase}, which operates over arbitrary window sizes and maintains a sample of input data that is exponentially weighted towards more recent points. Collectively, the 
databases also enable offline experimentation, \eg{} A/B testing. The \emph{System State DB} is used both by the policy engine 
to enable decision making. Finally \emph{Diagnostic Logs} provide a full history for system debugging.

\section{Progression Testing: Automatic Monitoring and Quality Control}
Monitoring and quality control have been cited as of critical importance for managing \ac{ML} systems in production \citep{breck2016mltestscore}. Here we outline our approach to ensuring that a high bar is met for both.

\subsection{Data Monitoring}
\label{sec:drift}
The key observation is that the world is non-stationary, whereas most of the \ac{ML} models used in deployment make the assumption (either explicitly or implicitly) that it is stationary. Hence there is a clear need to identify non-stationarity in its different forms in order to determine whether or not a trained \ac{ML} model is likely to behave as it did during the training process or not. There are many types of concept drift that have been identified \citep{quionero2009dataset}, each of which can affect model performance. 
We provide an overview of some approaches to dataset shift detection in \Cref{sec:sota_shift}.

At a minimum, the outputs of the data monitoring subsystem are: 
\begin{enumerate*}[label=\emph{\roman*)}]
  \item type of shift;
  \item shift magnitude;
  \item quantification of uncertainty.
\end{enumerate*}
It is also desirable that explanations should be given for the shift occurring, such as which feature/dimension was most important.
Solutions to the problem of dataset shift include domain adaptation or transfer learning (see \cite{pan2010survey}), or, more pragmatically, one can simply retrain the model. In nearly all practical cases the latter is simpler to implement. The problem can be lifted to warm-starting using \ac{HPO}, \eg{} by \acl{BO} \cite{snoek2012practical}, for which scalable methods exist whose complexity is linear in the number of observations \citep{perrone2017multiple}. 

\subsection{Prediction Monitoring}
\label{sec:prediction_monitoring}
The prediction monitoring subsystem maintains current estimate of the state of the ``world'' written to system state database. In practice this is a set of things like the output of the sketcher to estimate whether it's likely that the data distribution has shifted, plus contextual information such as the time since last retraining, current date/time, cost of retraining, value of predictions (\ie{} how important is success), estimated optimal policy, estimated value of different state/action pairs \etc{}. Note that the prediction monitoring can make use of the same dataset shift components described earlier, only this time operating on the level of predictions rather than input streams.

\section{Automating the \ac{ML} pipeline life-cycle}
When managing model retraining we face several difficult problems:
\begin{enumerate}[topsep=0pt,itemsep=-1ex,partopsep=1ex,parsep=1ex]
  \item {\bf Horizon}: How quickly does the most recent data-point need to become part of the model? How long does it take for data to become irrelevant? We need to find the right scope of data relevant for learning at the present moment. Generally, newer instances are more relevant, but in some cases (\eg{} retail), data from the previous quarter/year are more relevant.
  \item {\bf Cadence}: We need to know when to retrain. This requires an accurate estimate of the state of the world (\ie{} by analysing streams for possible anomalies and concept drift), and a set of possible actions (\eg{} retraining, using the existing model, model rollback, transfer learning). Given these we can learn the (locally) optimal policy: the best action to take given the current state. This decision involves balancing the operational cost against the potential downstream benefit resulting from  improvements to the model.
  \item {\bf Provenance}: It is essential that we are able to trace back from decisions to their underlying causes. 
  The system must include comprehensive logging as well as providing continuous health monitoring output streams.
  \item {\bf Costs}: What are the operational costs of retraining? What are the risks of not retraining?
\end{enumerate}

To this end, policy engine interacts with the trainer and predictor subsystems as follows.

\subsection{Trainer, \acl{HPO}, and Predictor}
\label{sec:trainer}
The trainer subsystem is a wrapper around the custom \ac{ML} model training algorithm to be used. Usually teams will have an existing \ac{ML} training pipeline. Here we would make use of that pipeline, at the simplest level only taking over the retraining step. It is important here that trainer stores any meta-data associated with training (\eg{} validation metrics, start time and duration, hyperparameter settings, trained model details) to the appropriate databases within the shared infrastructure (see \Cref{sec:cache}). The trainer can also benefit from \ac{HPO} to automatically warm-start the training of the \ac{ML} model.
%
As with the trainer subsystem, the predictor subsystem will likely be part of the existing pipeline. The main difference here is that this system is responsible for the deployment of newly trained models. 
The predictor subscribes to a continuous model stream coming from the trainer.

\subsection{Model Policy Engine}
\label{sec:policy}
The model policy engine must ensure that retraining is done at the optimal cadence \emph{for the given business metrics}. Every different use-case will have different metrics, but all will have to balance costs versus potential future benefits. The baseline is a simple set of rules, which for some simple cases will be sufficient. In general, it will be desirable for the policy itself to be learnable, where it would take the form of a \acl{RL} system \citep{sutton2011reinforcement}, where the state space is governed by the outputs of the data monitoring, trainer, and prediction monitoring subsystems, the action space 
is that of the policy engine, 
and the rewards are constructed based on the business metrics. Of course it would be expected that the system would make sub-optimal decisions during learning of the optimal policy. This is an active area of research.

\section{Conclusions}
\label{sec:conclusions}
We have presented a reference architecture for self-maintaining intelligent systems that can learn continually, as data arrives. 
This enables systems that effectively manage \ac{ML} models in production, by adapting to shifting data distributions, coping with outliers, retraining when necessary, and adapting to new tasks. We highlighted the main aspects of each component of the system, but note that each comprises a significant effort in terms of both engineering and science. Widespread adoption of such a system would of course be extremely challenging for teams with established production pipelines, but we feel strongly that this investment would pay dividends. Furthermore, the architecture has been designed to be entirely modular, such that components can be adopted in a stage-wise manner. Ultimately, we believe that the version of zero-touch \ac{ML} described here is a stepping stone towards fully continual learning \citep{silver2013lifelong}, an often cited desirable end-point for intelligent systems.




\newpage
\small

\bibliography{mlzero}

\begin{thebibliography}{10}

\bibitem{feurer2015efficient}
Matthias Feurer, Aaron Klein, Katharina Eggensperger, Jost Springenberg, Manuel
  Blum, and Frank Hutter.
\newblock Efficient and robust automated machine learning.
\newblock In C.~Cortes, N.~D. Lawrence, D.~D. Lee, M.~Sugiyama, and R.~Garnett,
  editors, {\em Advances in Neural Information Processing Systems 28}, pages
  2962--2970. Curran Associates, Inc., 2015.

\bibitem{fblearnerflow}
Jeffrey Dunn.
\newblock {Introducing FBLearner Flow: Facebook’s AI backbone}.
\newblock
  \url{https://code.fb.com/core-data/introducing-fblearner-flow-facebook-s-ai-backbone/},
  2016.
\newblock Accessed: 2018-10-19.

\bibitem{michelangelo}
Jeremy Hermann and Mike~Del Balso.
\newblock {Meet Michelangelo: Uber's Machine Learning Platform}.
\newblock \url{https://eng.uber.com/michelangelo/}, 2017.
\newblock Accessed: 2018-01-26.

\bibitem{bailis2017macrobase}
Peter Bailis, Edward Gan, Samuel Madden, Deepak Narayanan, Kexin Rong, and
  Sahaana Suri.
\newblock {MacroBase}: Prioritizing attention in fast data.
\newblock In {\em Proceedings of the 2017 ACM International Conference on
  Management of Data}, pages 541--556. ACM, 2017.

\bibitem{breck2016mltestscore}
Eric Breck, Shanqing Cai, Eric Nielsen, Michael Salib, and D.~Sculley.
\newblock What’s your {ML} test score? a rubric for {ML} production systems.
\newblock 2016.

\bibitem{quionero2009dataset}
Joaquin Quionero-Candela, Masashi Sugiyama, Anton Schwaighofer, and Neil~D
  Lawrence.
\newblock {\em Dataset shift in machine learning}.
\newblock The MIT Press, 2009.

\bibitem{pan2010survey}
Sinno~Jialin Pan and Qiang Yang.
\newblock A survey on transfer learning.
\newblock {\em IEEE Transactions on knowledge and data engineering},
  22(10):1345--1359, 2010.

\bibitem{snoek2012practical}
Jasper Snoek, Hugo Larochelle, and Ryan~P Adams.
\newblock Practical bayesian optimization of machine learning algorithms.
\newblock In {\em Advances in neural information processing systems}, pages
  2951--2959, 2012.

\bibitem{perrone2017multiple}
Valerio Perrone, Rodolphe Jenatton, Matthias Seeger, and Cedric Archambeau.
\newblock Multiple adaptive bayesian linear regression for scalable bayesian
  optimization with warm start.
\newblock {\em arXiv preprint arXiv:1712.02902}, 2017.

\bibitem{sutton2011reinforcement}
Richard~S Sutton and Andrew~G Barto.
\newblock Reinforcement learning: An introduction.
\newblock 2011.

\bibitem{silver2013lifelong}
Daniel~L Silver, Qiang Yang, and Lianghao Li.
\newblock Lifelong machine learning systems: Beyond learning algorithms.
\newblock In {\em AAAI Spring Symposium: Lifelong Machine Learning}, volume~13,
  page~05, 2013.

\bibitem{cormode2012synopses}
Graham Cormode, Minos Garofalakis, Peter~J. Haas, and Chris Jermaine.
\newblock Synopses for massive data: Samples, histograms, wavelets, sketches.
\newblock {\em Found. Trends databases}, 4(1-3):1--294, jan 2012.

\bibitem{bloom1970filter}
Burton~H. Bloom.
\newblock Space/time trade-offs in hash coding with allowable errors.
\newblock {\em Commun. ACM}, 13(7):422--426, July 1970.

\bibitem{cormode2005countmin}
Graham Cormode and S.~Muthukrishnan.
\newblock An improved data stream summary: the count-min sketch and its
  applications.
\newblock {\em Journal of Algorithms}, 55(1):58 -- 75, 2005.

\bibitem{flajolet2007hyperloglog}
Philippe Flajolet, {\'E}ric Fusy, Olivier Gandouet, and Fr{\'e}d{\'e}ric
  Meunier.
\newblock Hyperloglog: the analysis of a near-optimal cardinality estimation
  algorithm.
\newblock In {\em AofA: Analysis of Algorithms}, pages 137--156. Discrete
  Mathematics and Theoretical Computer Science, 2007.

\bibitem{durand2003loglog}
Marianne Durand and Philippe Flajolet.
\newblock Loglog counting of large cardinalities.
\newblock In {\em European Symposium on Algorithms}, pages 605--617. Springer,
  2003.

\bibitem{heule2013hyperloglog}
Stefan Heule, Marc Nunkesser, and Alex Hall.
\newblock Hyperloglog in practice: Algorithmic engineering of a state of the
  art cardinality estimation algorithm.
\newblock In {\em Proceedings of the EDBT 2013 Conference}, Genoa, Italy, 2013.

\bibitem{xiao2017better}
Qingjun Xiao, You Zhou, and Shigang Chen.
\newblock Better with fewer bits: Improving the performance of cardinality
  estimation of large data streams.
\newblock In {\em INFOCOM 2017-IEEE Conference on Computer Communications,
  IEEE}, pages 1--9. IEEE, 2017.

\bibitem{metwally2005efficient}
Ahmed Metwally, Divyakant Agrawal, and Amr El~Abbadi.
\newblock Efficient computation of frequent and top-k elements in data streams.
\newblock In {\em International Conference on Database Theory}, pages 398--412.
  Springer, 2005.

\bibitem{dunning2014computing}
Ted Dunning and OTMAR Ertl.
\newblock Computing extremely accurate quantiles using t-digests.
\newblock {\em github. com}, 2014.

\bibitem{shrivastava2004medians}
Nisheeth Shrivastava, Chiranjeeb Buragohain, Divyakant Agrawal, and Subhash
  Suri.
\newblock Medians and beyond: new aggregation techniques for sensor networks.
\newblock In {\em Proceedings of the 2nd international conference on Embedded
  networked sensor systems}, pages 239--249. ACM, 2004.

\bibitem{dasgupta2000experiments}
Sanjoy Dasgupta.
\newblock Experiments with random projection.
\newblock In {\em Proceedings of the Sixteenth conference on Uncertainty in
  artificial intelligence}, pages 143--151. Morgan Kaufmann Publishers Inc.,
  2000.

\bibitem{achlioptas2003database}
Dimitris Achlioptas.
\newblock Database-friendly random projections: Johnson-lindenstrauss with
  binary coins.
\newblock {\em Journal of computer and System Sciences}, 66(4):671--687, 2003.

\bibitem{agarwal2013mergeable}
Pankaj~K Agarwal, Graham Cormode, Zengfeng Huang, Jeff~M Phillips, Zhewei Wei,
  and Ke~Yi.
\newblock Mergeable summaries.
\newblock {\em ACM Transactions on Database Systems (TODS)}, 38(4):26, 2013.

\bibitem{scholkopf2000support}
Bernhard Sch{\"o}lkopf, Robert~C Williamson, Alex~J Smola, John Shawe-Taylor,
  and John~C Platt.
\newblock Support vector method for novelty detection.
\newblock In {\em Advances in neural information processing systems}, pages
  582--588, 2000.

\bibitem{guha2016robust}
Sudipto Guha, Nina Mishra, Gourav Roy, and Okke Schrijvers.
\newblock Robust random cut forest based anomaly detection on streams.
\newblock In {\em International Conference on Machine Learning}, pages
  2712--2721, 2016.

\bibitem{aminikhanghahi2017survey}
Samaneh Aminikhanghahi and Diane~J Cook.
\newblock A survey of methods for time series change point detection.
\newblock {\em Knowledge and information systems}, 51(2):339--367, 2017.

\bibitem{adams2007bayesian}
Ryan~Prescott Adams and David~JC MacKay.
\newblock Bayesian online changepoint detection.
\newblock {\em arXiv preprint arXiv:0710.3742}, 2007.

\bibitem{kawahara2012sequential}
Yoshinobu Kawahara and Masashi Sugiyama.
\newblock Sequential change-point detection based on direct density-ratio
  estimation.
\newblock {\em Statistical Analysis and Data Mining: The ASA Data Science
  Journal}, 5(2):114--127, 2012.

\bibitem{liu2013change}
Song Liu, Makoto Yamada, Nigel Collier, and Masashi Sugiyama.
\newblock Change-point detection in time-series data by relative density-ratio
  estimation.
\newblock {\em Neural Networks}, 43:72--83, 2013.

\bibitem{baena2006early}
Manuel Baena-Garc{\'\i}a, Jos{\'e} del Campo-{\'A}vila, Ra{\'u}l Fidalgo,
  Albert Bifet, Ricard Gavald{\`a}, and Rafael Morales-Bueno.
\newblock Early drift detection method.
\newblock 2006.

\end{thebibliography}

\newpage
\begin{appendices}
\crefalias{section}{appsec}
\crefalias{subsection}{appsec}
\section{Review of the \ac{SOTA}: Sketcher/Compressor}\label{sec:sota_sketch}
\vspace{-1ex}
For a review of sketching algorithms see \cite{cormode2012synopses}.

\paragraph{Bloom Filters.} 
A Bloom filter \citep{bloom1970filter} is a fixed-size set data structure that is optimised for membership queries. Membership questions that can be asked are:
\begin{enumerate*}[label=(\roman*)]
  \item The item has definitely not been stored;
  \item The item has probably been stored.
\end{enumerate*}
The membership queries have a small, controllable, false-positive probability. Bloom filters will never return false negatives. You can never ask a Bloom Filter for the set of elements it contains. The basic idea is to use multiple hash functions to bit values, and only return true if all hash functions return 1. Adding extra hash functions reduces chances of false positives, but increases the chance of collisions. If $n$ items are being stored in a Bloom filter of size $m$, and $k$ hash functions are used, then the chance of a membership query false positive is $\approx (1 - e^{-kn/m})^k$. The optimum choice for $k$ is then $(m/n)\log 2$. As an example, a common setting is for $m = 10n$, $k = 7$, which would give a false positive rate of $< 1 \%$.

\paragraph{Count-min Sketch.} 
For estimating the counts of items in data streams, the count-min sketch \citep{cormode2005countmin} is a sublinear space data structure which allows fundamental queries in data stream summarisation such as point, range, and inner product queries to be approximately answered very quickly. It can also be applied to finding quantiles, frequent items, or tracking which items exceed a given popularity threshold. Large counts preserved fairly accurately, but small counts may incur greater relative error. 
For sketch of size $s$, the error is proportional to $1/s$ (compared to $1/\sqrt{s}$ for random sampling). The size of the data structure can be considered to be independent of the input size, depending instead on the desired accuracy guarantee only. For target accuracy of $\epsilon$, one can simply fix $s \propto 1/\epsilon$ that does not vary over the course of processing data. It is also worth mentioning that range queries can be computed using arrays of count-min sketches.

\paragraph{HyperLogLog.} 
If the only quantity of interest is the number of unique/distinct items (cardinality estimation), the HyperLogLog algorithm  \citep{flajolet2007hyperloglog} and its derivatives compute this, where the cost only depends on log of log of quantity computed. HyperLogLog is an extension of the original LogLog algorithm \citep{durand2003loglog}. The basis of these algorithms is the observation that the cardinality of a multiset of uniformly distributed random numbers can be estimated by calculating the maximum number of leading zeros in the binary representation of each number in the set. 
More memory efficient versions of HyperLogLog have been proposed. A version called HyperLogLog++ \citep{heule2013hyperloglog} uses 64-bit hash functions instead of 32-bit, which reduces the probability of collisions. HLL-TailCut+ \citep{xiao2017better} achieves estimation standard error $1.0/\sqrt{m}$ using memory units of three bits each (rather than the five-bit memory units used by HyperLogLog). This makes it possible to reduce the memory cost of HyperLogLog by $45\%$: \eg{} when the target error is $1.1\%$, HyperLogLog needs $5.6$ kB of memory, whereas HLL-TailCut+ needs $3$ kB to attain the same accuracy.

\paragraph{Stream-Summary.} 
For computing the so-called ``heavy-hitters'', \ie{} the $k$ most frequent items, the stream summary structure \citep{metwally2005efficient} can be used.  Stream-Summary traces a fixed number of elements that are with high probability the most frequent ones. If one of these elements occurs, the corresponding counter is increased. If a new non-traced element appears, it replaces the least frequent traced element. Querying the most frequent elements and corresponding frequencies from the data structure is trivial, but to answer whether these estimates are exact (guaranteed) or not is much less so; corresponding algorithms are described in \cite{metwally2005efficient}.

\paragraph{t-digest.} 
For rank-based statistics, such as quantiles and trimmed means, a new data structure ``t-digest'' for accurate on-line accumulation has been proposed \cite{dunning2014computing}. The t-digest construction algorithm uses a variant of 1-dimensional k-means clustering to produce a data structure that is related to the Q-digest \citep{shrivastava2004medians}, except that the t-digest can handle floating point values while the Q-digest is limited to integers. The accuracy of quantile estimates produced by t-digests can be orders of magnitude more accurate than those produced by Q-digests in spite of the fact that t-digests are more compact when stored on disk. The t-digest algorithm is also readily parallelisable, making it useful in map-reduce and parallel streaming applications.

\paragraph{Random projections.} 
If the input streams are of the form of high-dimensional numerical data, we can seek to reduce the dimensionality while preserving fidelity of the data.  \ac{PCA} aims to extract a small number of ``directions'' from the data whilst capturing most of the variation of the inputs. However \ac{PCA} requires finding eigenvectors of the covariance matrix, which rapidly becomes unsustainable for large matrices. An alternative approach of random projections shows that it suffices to use (a slightly larger number of) random vectors \citep{dasgupta2000experiments,achlioptas2003database}. The random projection of each row of the data matrix can be seen as an example of a sketch of the data (and indeed the Count-Min sketch can be viewed as a random projection).

In practice, it may be hard to know ahead of time which queries will be required. One option is to compute a range of sketches simultaneously with a specified error rate (\eg{} $4\%$), which would still represent a significant saving \footnote{For a stream of $10^7$ 32-bit integers with $10^6$ distinct values, a Bloom filter, count-min sketch, stream-summary and LogLog counter combined would require $<0.7$ Mb, versus 40 Mb for the raw data}. One topic not covered here is that of mergeable summaries (see \eg{} \cite{agarwal2013mergeable}). These are important if the summaries must be distributed, either for computational or systems architectural reasons.

\section{Review of the \ac{SOTA}: Dataset Shift Detection}
\label{sec:sota_shift}
Given a source and target datasets $D_S$ and $D_T$ consisting of inputs $x$ and targets $y$, drawn from underlying distributions $p_S(x,y)$ and $p_T(x, y)$ respectively, we are interested in detecting different types of shift between $p_S$ and $p_T$. An additional assumption will be that $S$ remains fixed until the detection of drift, while $T$ is produce by a sliding window on the incoming data streams. Types of shift include: 
\begin{enumerate*}[label=(\roman*)]
  \item simple covariate shift: only the distributions of covariates $p_T(x)$ change and everything else is the same;
  \item prior probability shift: only $p_T(y)$ changes and everything else stays the same;
  \item sample selection bias: the distributions differ as a result of an unknown sample rejection process;
  \item imbalanced data is a form of deliberate dataset shift for computational or modelling convenience;
  \item domain shift: changes in measurement;
  \item source component shift: changes in strength of contributing components;
  \item anomaly detection: transient shifts, normally in $p_T(x)$.
\end{enumerate*}

Different approaches to detecting shift that are appropriate for the different kinds of shift. These fall into two categories: supervised and unsupervised. Supervised approaches include examining the progressive error of a model (\eg{} a classifier or a regressor), or maintaining an external holdout dataset. Whilst these are attractive for their simplicity, they rely on having labelled data available, or a mechanism for keeping the holdout dataset up-to-date. These approaches may make sense for examining the behaviour of trained models during operation. In most cases however, we will have to resort to unsupervised methods. The simplest and most scalable approaches are based on statistical distances between $p_S$ and $p_T$, such as the \ac{PSI}, or the Kolmogorov-Smirnov statistic, but these are often limited to low dimensional or real-valued data. A natural measure of divergence between distributions is the \ac{KL} divergence, although for general use this requires probabilistic assumptions to be made. A popular non-parametric method is the histogram intersection method, which can be used on both real and categorical data, but is very sensitive to the number of bins used in the histogram calculation, and also suffers in high dimensions.

For the problem of detecting anomalies (abrupt but transient shifts), early approaches that scale well to higher dimensions include the one class \ac{SVM} \citep{scholkopf2000support}, which are flexible but are computationally prohibitive. A more efficient approach based on a random cut forest \citep{guha2016robust} computes a sketch of the data upon which the anomaly detection decisions are made (this sketch can in fact be included as part of the sampler/sketcher component and re-purposed for other uses). 

If the shifts are abrupt but non-transient, then the problem is usually called change-point detection (for a recent survey see \cite{aminikhanghahi2017survey}), with a divide between parametric methods that build a model of $p_S(x)$ \citep{adams2007bayesian} and non-parametric approaches based on estimating the ratio between the densities $p_S(x)$ and $p_T(x)$. The rationale of density ratio estimation is that knowing the two densities implies knowing the density ratio, but that the inverse is not true since the decomposition is not unique. As a result, direct density ratio estimation is substantially simpler than density estimation. An example of this is \ac{KLIEP} \citep{kawahara2012sequential}, which estimates the density ratio using KL divergence, and there have since been modifications to this method which improve scalability \citep{liu2013change}.

Gradual drifts are the most challenging type to deal with. There have been some attempts to tackle this, such as \ac{EDDM} \citep{baena2006early}, but this method requires waiting for a minimum of 30 classification errors before calculating the monitoring statistic at each decision point, which may be too high for many real applications. A general strategy will be to ensure that the overall performance of the system is as is expected through the use of health monitoring streams.
\end{appendices}
\end{document}